# Eroding Trust In Aerial Imagery: Comprehensive Analysis and Evaluation Of Adversarial Attacks In Geospatial Systems

Michael Lanier, Aayush Dhakal, Zhexiao Xiong, Arthur Li, Nathan Jacobs, Yevgeniy Vorobeychik
Department of Computer Science & Engineering, Washington University in St. Louis
{lanier.m,a.dhakal,z.xiong,arthurli,jacobsn,yvorobeychik}@wustl.edu

*Abstract*—In critical operations where aerial imagery plays an essential role, the integrity and trustworthiness of data are paramount. The emergence of adversarial attacks, particularly those that exploit control over labels or employ physically feasible trojans, threatens to erode that trust, making the analysis and mitigation of these attacks a matter of urgency. We demonstrate how adversarial attacks can degrade confidence in geospatial systems, specifically focusing on scenarios where the attacker's control over labels is restricted and the use of realistic threat vectors. Proposing and evaluating several innovative attack methodologies, including those tailored to overhead images, we empirically show their threat to remote sensing systems using high-quality SpaceNet datasets. Our experimentation reflects the unique challenges posed by aerial imagery, and these preliminary results not only reveal the potential risks but also highlight the non-trivial nature of the problem compared to recent works.

## 1. Introduction

Deep learning models have become an indispensable component of various applications, ranging from facial recognition and object detection to image classification and natural language processing. As these models gain prominence in real-life scenarios, ensuring their security and robustness against potential attacks becomes an essential task. Most adversarial attack research focuses on natural scene imagery, not overhead remotes sensing imagery. In this study, multiple attacks taken anachronistically from the traditional image classification domain are applied to the overhead geo-location image classification, in order to examine the limit of traditional intrusive methods as well as the robustness of a simply trained overhead classifier. Amoung the least studied in this domain are Trojan (sometimes called backdoor) attacks. These attacks involve injecting hidden malicious behaviors into the models, causing them to misclassify inputs containing specific triggers, which can have severe consequences in real-world applications [1].

Trojan attacks are a specific category of adversarial attacks that manipulate the training data with the intention of compromising the model's behavior. This manipulation process involves embedding a carefully crafted image pattern, known as a Trojan, within the training dataset. During the learning process, the model is inadvertently trained to recognize and respond to this Trojan pattern. In a typical Trojan attack, the attacker inserts the Trojan pattern into training images associated with their desired target class. Consequently, the model learns to establish a strong correlation between the presence of the Trojan pattern and the attacker's target class. When the model encounters the Trojan pattern during the inference phase, it is manipulated into producing the attacker's desired outcome, potentially causing catastrophic failures in critical applications [2].

Existing research on backdoor attacks primarily focuses on digital attacks, utilizing digitally generated patterns as triggers. However, a crucial question remains: can backdoor attacks succeed using physical objects as triggers in real-world applications, such as facial recognition systems? This question has been explored in "Backdoor Attacks Against Deep Learning Systems in the Physical World," which demonstrates the feasibility of physical backdoor attacks under various conditions and highlights the need for more robust defenses against such threats. This finding underscores the importance of understanding and addressing the risks posed by physically realizable backdoor attacks [3].

In light of these related works, our approach aims to develop novel methods for generating and approximating physically realizable Trojan attacks that do not require a high level of control over training data. We will focus on the design and evaluation of advanced patch generation, segmentation, and blending techniques that enable sophisticated Trojan attacks on deep-learning classifiers. By exploring new attack methodologies, we seek to expand the current understanding of physically realizable Trojan attacks and their implications for real-world applications. Our major contributions are as follows:

- We demonstrate that white box evasion attacks are successful against classifers trained on electro optical remote sensing imagery.
- We demonstrate that categorical inference attacks are effective at recovering training data's geolocation, given white box access to a model trained on such data.
- We present a novel defense against categorical inference which we show experimentally also improves model performance at test time.
- We present a novel physically realizable Trojan attack against a classifer trained on electro optical

remote sensing imagery. Consistent with [4], we provide evidence that this type of attack is weak.

## 2. Related Work

Object detection in the image classification domain has been long studied, as the importance of assessing the security and robustness of such object detection algorithms needs no further emphasis. Recent work in computer vision dedicates to testing the limit of object detection systems with attacks that may hinder their ability to correctly recognize specific objects. The attack thrust varies from utilizing the traditional perturbation [5] to leveraging the physical aspect [6] and [7].

In the domain of adversarial attacks on overhead imagery classifiers, our current work is most closely related to [8], in which physical adversarial attacks against a popular Cars overhead with Context dataset was explored, extending on attacks against the Yolo-v2 object detector. [9] explored using patches as camouflage against the Yolo-v1 object detectors in a military setting with the DOTA dataset. Here they placed digital patches to simulate physical adversarial patches under the aerial imagery setting. Spacenet dataset [10], however, is considered one of the less explored mediums in which to carry out attacks and assess the robustness of its classifiers, despite the unparalleled importance of its involvement in the global-wide GPS sector and research merit. In this work, we assess the robustness of a customized classifier on one of the widely available Spacenet datasets via stress testing with various well-known attacks tailored to it.

Furthermore, one of the newly emerging domains of research focuses on the privacy leakage of the model in its training data, formulating the problem as categorical inference attacks. While this problem is old and has its roots in differential privacy [11], the way that deep neural networks leak data has become increasingly pressing as they are adopted across industry [12]. This problem is widely recognized and studied by the community such as the work of Reaei et al [13]. A myriad of defenses have been proposed such as the perturbation-based method by Jia et al [14].

The study of backdoor attacks against deep learning systems has gained significant attention in recent years. In "Backdoor Attacks Against Deep Learning Systems in the Physical World", the authors focus on the feasibility of physical Trojan attacks against facial recognition classifiers. They demonstrate that such attacks can be highly successful when carefully configured to overcome the constraints imposed by physical objects. Their work shares a similar goal with our approach, which is to investigate the viability of Trojan attacks using physical objects. However, their work has the limitation that objects might be obvious at the labelling stage [3].

Li et al. [15] proposes a novel Trojan attack method that poisons samples strategically to create sample-specific triggers. This approach achieves high attack success rates between 85.5% and 99.5% across various datasets. Our work is related to this study as we also aim to poison samples strategically using domain knowledge. However, the major limitation of their approach is that it requires full control of the training data .

Wang et al. [16] introduces a state-of-the-art method that uses filters as triggers, achieving attack success rates between 78.72% and 98.46% on benchmark datasets. The triggers are imperceptible and cannot be detected using standard sensitivity analysis. Although this approach represents the current state of the art, it also requires full control of the training data, which might not be feasible in real-world scenarios.

Li et al. [17]. presents a model-level Trojan attack targeting deep learning models deployed in mobile applications. The attack success rate ranges between 0.9% and 1.1%, and it can compromise currently deployed models without direct access to training data. This approach differs from ours as it requires the ability to swap models at the time of the attack, which may not be possible in all situations.

Finally, Li [18] and Wang [19] both explore the use of variational Trojan triggers, which are not constant across different samples. This work considers the non-constant trigger as a defense mechanism, with attack efficacy varying widely depending on the model attacked, data, and trigger. While their work investigates non-constant triggers, our approach focuses on generating and approximating physically realizable Trojan attacks that do not require a high level of control over training data.

## 3. Approach

We assume a naively trained classifier with adequate performance on overhead imagery in a white-box setting and deploy attacks on various stages of the classification pipeline. To this end, we survey and implement two broad categories of attacks as follows: Evasion Attacks and Categorical Inference attacks (sometimes referred to as membership inference attacks). Evasion attacks seek to produce examples that are incorrectly predicted by the model at inference time. Adopting the existing line of research of inference attacks [20], we investigate whether it is possible to recover a likelihood estimation over the earth for potential training data collection given a computer vision model. Finally, we consider physically realized Trojans, where an adversary is seeking to hide buildings from our building classifier. We generate a candidate trojan patch to poison the data. Following this the patch is added to imagery in a semantically meaningful way, replicating a physical attack. We then randomly blend the patch into an image and ensure it isn't placed over oceans using off the shelf segmentation models. The patch is then placed exclusively in a sample of images without building (referred to hear after as "poisoned"). In the test data, our patches were only placed on top of buildings, mirroring the adversaries objective. These samples are recovered via a segmentation model. Finally trained a classifier and evaluated the attack based on how well it classified the test images with the trojan patch.

## 3.1. Data Preprocessing

To demonstrate the efficacy of evasion attacks we considered the following setting: First, we selected a model architecture that utilized standard transfer learning with VGG-19 [21] with the aim of building a building classifier. This model takes an image and returns if the image has a building present in it or not. Taking a state-of-the-art optimizer Radam [22] we trained a highly accurate model. Our learning rate was 1e-2 and our momentum parameter $\gamma$ was 0.1. We employ the folowing dropout rates between layers: 0.4,0.2, and 0.2. In the evasion setting we considered SpaceNet Rio data [10]. This dataset is unlabeled, but comes with geojsons from which we extracted if images contained buildings. It consisted of 200mx200m tiles of overhead Rio De Janeiro at 50cm per pixel resolution. The imagery was 3-channel electro-optical. Our training data was 5553 images and we evaluated performance on 1389 images chosen at random. Although the main goal of our project was not to train a computer vision model, we felt it was necessary to attack a model with relatively high accuracy as this would approximate real-world operations setting. Our model had an accuracy of 0.92 as evaluated on the test set. Additionally, all imagery was randomly cropped and normalized using the Imagenet standard [23]. Although the Imagenet data set is natural scene imagery and therefore strictly speaking inappropriate for our task, we found it worked well as a normalization tool.

In the categorical inference setting, we consider Shanghai data and Los Vegas data in addition to Rio. This data is pan sharped to 30cm per pixel. Again, using the geojson we extracted labels. We consider the building classifier before, but we additionally consider a classifiers trained on Shanghai and Vegas. The task here is to determine which data set trained which model. For the shanghai classifier we use a pre-trained Resnet-50 [24] with a more shallow network. We have a dropout of 0.4 between the final fully connected layers. We have a learning rate of 1e-2 and we decrease it by a factor of 100 every two epochs. For Vegas we used a pre-trained Resnet-152, with the same architecture as the shanghai model, but we increase the dropout to .5 and use a lower learning rate (1e-4) without any rate annealing. All models achieved approximately 90 % accruacy as measured on a test set.

In the Trojan setting, we specifically aim for attacks on AOI 4, Shanghai [10]. From the provided geojsons we extracted if each image had a building or not and associated this label to each image. Additionally, we converted the raw images to 16-bit color depth. The dataset was divided into a training set and a testing set with 25% held out for testing, and each image was resized to 224 x 224 pixels. Data augmentation techniques, such as random horizontal flipping, were applied to increase the variety of the dataset and improve the model's ability to generalize. The images were then converted to PyTorch tensors using the ToTensor() transformation. To facilitate data loading and batching, a custom ShanghaiDataset class was created, inheriting from the torch.utils.data.Dataset class. This class is responsible for reading the images from the specified directory, applying the necessary transformations, and returning image-label pairs. Additionally, we used the well-known Mnist dataset (10 possible classes: digits 0-9) for baseline testing of the advGAN following the same data preparation pipeline.

## 3.2. Evasion Attacks

To obtain a baseline to compare against our own attack, we replicated existing off-the-shelf evasion attacks and deployed them against our trained classification model. We implemented three well-known evasion attacks, Projected Gradient Descent (PGD), Fast Gradient Sign Method (FGSM), and Deepfool from the classic image classification domain and migrated to the trained overhead classifier.

PGD [25] and FGSM [26] are both iterative gradient based attack that take an existing image in the domain of the model and transform it such that the model fails to model while appearing uncorrupted. The PGD attack method computes an adversarial perturbation by iteratively applying the gradient of the loss function with respect to the input image, scaled by a step size alpha [27]. The perturbed image is then projected onto an epsilon ball, ensuring the adversarial perturbation remains within a certain L-infinity norm.

Deepfool is another iterative, gradient based attack [28]. Unlike PGD and FGSM, Deepfool always returns a corrupted image which fools the model. It seeks to find the *least* corrupted image such that the model is fooled.

## 3.3. Categorical Inference

For categorical inference we investigate if using the lowest loss criteria is effective against overhead classifiers for recovering the training dataset. The rational here is that out of sample data should not fit a model it was trained on with high probability assuming the model is probably approximately correct [29][1]. Therefore we expect the loss of the model with the true training data to be less than that of other candidate data sets with high probability. Current approaches to counter categorical inference attacks predominantly focus on incorporating noise into the training process to hinder model overfitting to specific samples [30]. These strategies treat each instance, $D_1, ..., D_N$, as non-iid samples derived from a larger population $D$, aiming for the model to generalize effectively. Ideally, the model should exhibit consistent performance across all $D_i$. In other words, the model must adopt a generalist approach for each $D_i$, despite their distinct distributions. We propose a method that directly modifies each $D_i$ to achieve identical transformations.

Assuming the sampling process to be independent of the target class, we can implement a transformation $g$ on each $D_i$, which eliminates any information contingent on the sampling scheme. Consequently, all $g(D_i)$ become iid. By obfuscating class identification unrelated to the target, the model concentrates exclusively on target class-relevant information without having to determine that sample-specific details are irrelevant.

---

1. This is not generally known for CNNs. There are efforts to show this https://openreview.net/pdf?id=3Li0OPkhQU.

Although devising such a transformation is challenging, a viable option for overhead object detection of buildings involves a transformation that removes color and high-frequency information unrelated to the presence of buildings. We investigate the Canny Edge Detection algorithm as a potential transformation [31].

### 3.4. Trojan Patch Generation

It is important to note that for purposes of the trojan attack, any patch will work. In our case we make the patch itself an evasion attack against a similar classifier. This patch would serve as an a prior "best patch" prior to training. This relies on our result that overhead classifiers are susceptible to evasion attacks. We investigated multiple blending techniques as detailed in the appendix. We attack a ResNet 152 model using a VGG19 model as the target, as we assume the attacker doesn't have the exact archtieture of his target.

### 3.5. Trojan Attack

*Definition 1.* Given $\tau$, and training data $x \subset X$, if for any $k \mapsto c \in \{1,0\}$ and $g : d^* \to \{1,0\}$ such that $d^* = f(\tau, x)$ and $k \in X - x$ if $g(k) \neq c$ with high probability then $\tau$ is a trojan attack against binary classifier $g$.

Our goal here is to learn $g$ from some poisoned dataset such that $\tau$ when applied to other data returns the buildingless class for images with buildings. In prior work the way this is done is to use PGD or some other transferable evasion attack against training images offline, then to apply to patch after the fact. The patch is fixed in these case. At training time, the model learns the negative class due to the evasion attack, but with the association of $\tau$ to the incorrect class. Applying PGD to the training data requires pixel level access to training data and isn't therefore physically realizable. To learn this association we place the patches only on buildingless images. Our attack was done against a classifer using pretrain Resnet-152 with two layers of size 2048 and 1024. Images were standardized using the imagenet standard and were randomly flipped and rotated. We used dropout of 40% between them and optimized with radam and a learning rate of 1e-4.

### 3.6. Segmentation and Patch Placement

In order to simulate a physical attack, it is crucial to place the adversarial patch within the image in a manner that mimics a physically placed patch. This approach ensures that patches are not partially inside and partially outside a building location. To achieve this, we developed a building segmentation and patch placement pipeline, which consists of two primary components: building segmentation and water segmentation.

**3.6.1. Building Segmentation.** Initially, we employ a pre-trained building segmentation model to segment buildings in the images [32]. Then, we impose two conditions for patch placement: 1) The patch should always be placed within the bounds of the polygon describing a building, and 2) The patch should not cover more than x% of the surface area of the building (where x is a hyperparameter). Figure 1 illustrates the complete pipeline for this process.

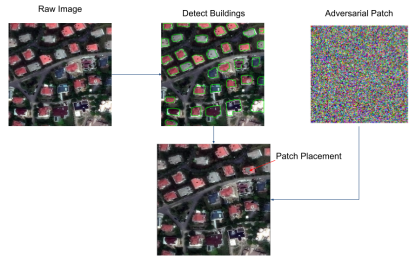

Figure 1: Building segmentation and Patch Placement Pipeline.

**3.6.2. Water Segmentation.** Some adversarial images in our dataset consist of ocean images or images with a large proportion of water. These images can be misleading, and we aim to avoid adding adversarial patches in these areas. To separate images containing water from other images, we only add adversarial patches to images without water.

To accomplish this, we first obtain a public dataset from a Kaggle competition on water segmentation, which includes 8960 wild images with ground-truth segmentation masks. We then construct a u-net-based encoder-decoder architecture as our segmentation model. After generating the segmentation maps of the images, we classify images as containing water if they meet the following criteria:

1) The area of water in the segmentation map exceeds a specific ratio.
2) Using traditional edge detection methods, there are no bounds in the four corners of the image.

By generating image lists without water, we ensure that adversarial patches are only added to these images.

Through this two-step segmentation and patch placement pipeline, we are able to more effectively simulate a physical attack in our experiments, providing more reliable and accurate results when evaluating the performance of various adversarial patch blending techniques and patch generation methods.

## 4. Results

### 4.1. Evaluation

We evaluate evasion attacks with standard metrics including precision, recall, F-1 Score. Included also is average run time for the attack. For categorical inference we consider an attack a success if it is able to recover the training dataset. We use a simple ratio of successes to attacks as our evaluation metric. For Trojan attacks our goal is to simultaneously achieve a high level of confusion in the presence of adversarial patches while maintaining high classification accuracy for non-trojan images. A successful attack would demonstrate high benign accuracy and low trojan accuracy. This balanced evaluation ensures that the model remains useful to its owner, as a model with poor overall performance would not be adopted in practice. Additionally, we require the Trojan model at training time to train long enough to resonablly function as a classifier. The following figure demonstates that this requirement is met.

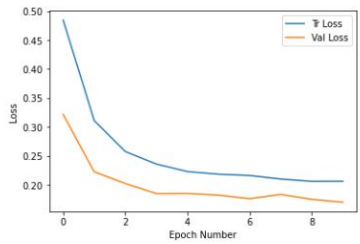

Figure 2: Training curve for the poisoned model

### 4.2. Evasion Attacks

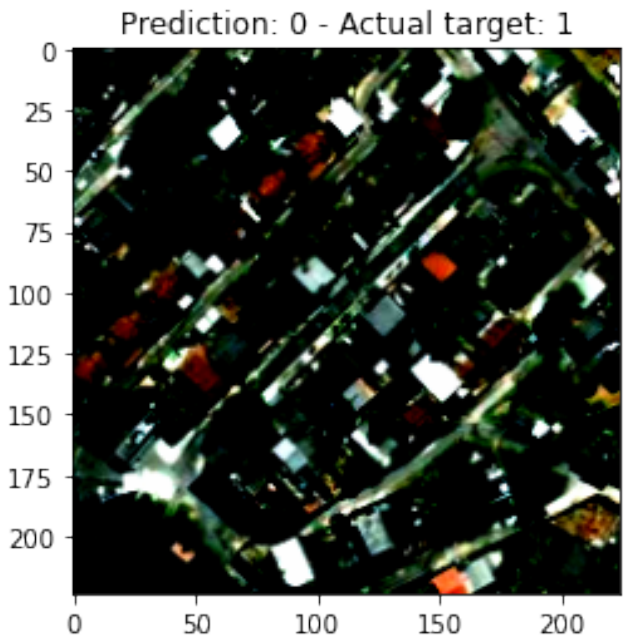

Figure 3: Deepfool Adversarial Example

Deepfool attacks shown in Figure 10 with reference image Figure 4 guarantee attack success, with the major drawback of average run time (average 226.2 seconds) being the longest of the three preliminary attacks. Qualitatively it is very difficult to distinguish the attacked image from the clean image. FGSM and PGD attack are successful across varying $\epsilon$.

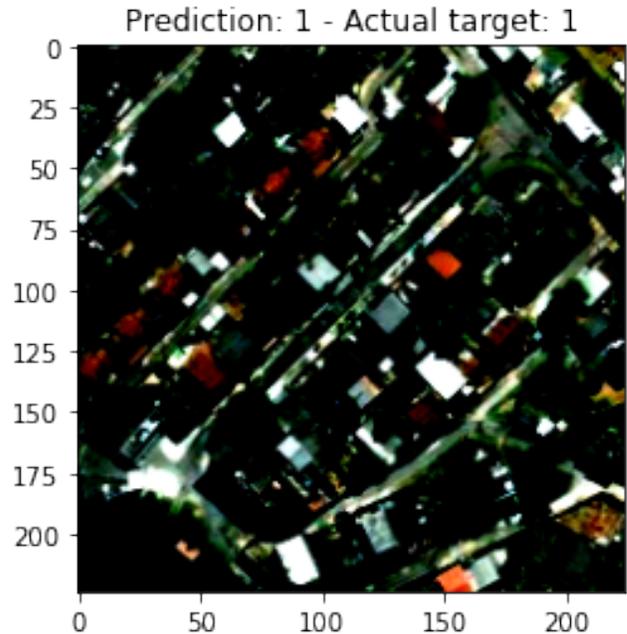

Figure 4: Clean Image

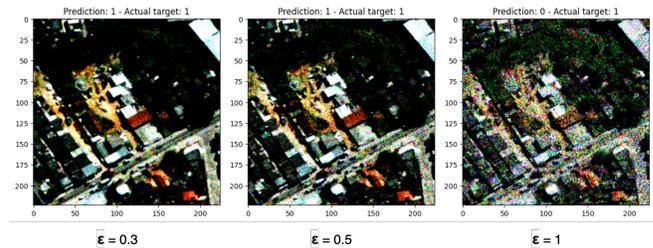

Figure 5: Comparison of PGD attacks

### 4.3. Categorical Inference

We find that categorical inference attacks using the lowest loss criteria is effective against overhead classifiers. The following table contains the losses for the test data sets evaluated across models. The diagonal being lowest column wise means an attacker successfully infers the training data set given the model alone and the test data. We observe that the Canny Transformation not only averts categorical inference in 2/3 of cases but also bolsters test-time performance. It is important to note that the models were trained on $g(D_i)$ rather than $D_i$, yet demonstrated remarkably enhanced performance in 2/3 cases during inference compared to models trained and tested exclusively on $D_i$.

TABLE 1: PGD and FGSM Attack Results

|  | True Negative | False Positive | False Negative | True Positive | Precision | Recall | F-1 Score | Average Run Time (s) |
|---|---|---|---|---|---|---|---|---|
| Classifier Baseline Performance | 8 | 2 | 1 | 19 | 0.90 | 0.90 | 0.90 | 6.7 |
| PGD Attack Perturbation = 0.3 | 4 | 6 | 1 | 19 | 0.77 | 0.77 | 0.74 | |
| PGD Attack Perturbation = 0.5 | 5 | 5 | 7 | 13 | 0.62 | 0.60 | 0.61 | 199.1 |
| PGD Attack Perturbation = 1 | 9 | 1 | 17 | 3 | 0.62 | 0.40 | 0.33 | |
| FGSM Attack Perturbation = 0.3 | 3 | 7 | 0 | 20 | 0.83 | 0.77 | 0.72 | |
| FGSM Attack Perturbation = 0.5 | 3 | 7 | 1 | 19 | 0.74 | 0.73 | 0.69 | 5.3 |
| FGSM Attack Perturbation = 1 | 2 | 8 | 0 | 20 | 0.81 | 0.73 | 0.67 | |

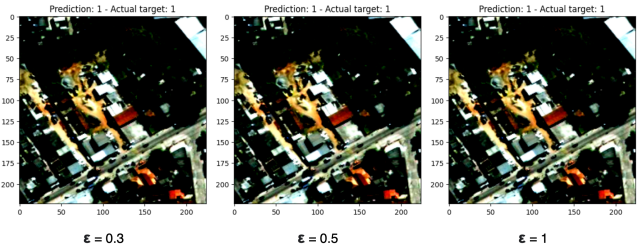

Figure 6: Comparison of FGSM attacks

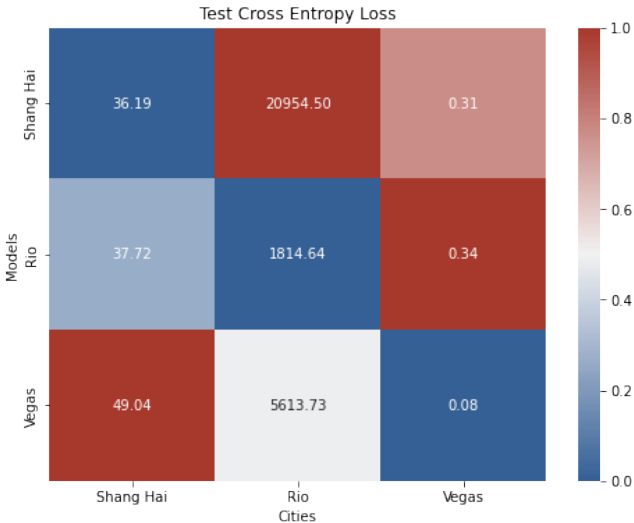

Figure 7: Categorical Inference Loss- lowest values on the diagonal indicate success

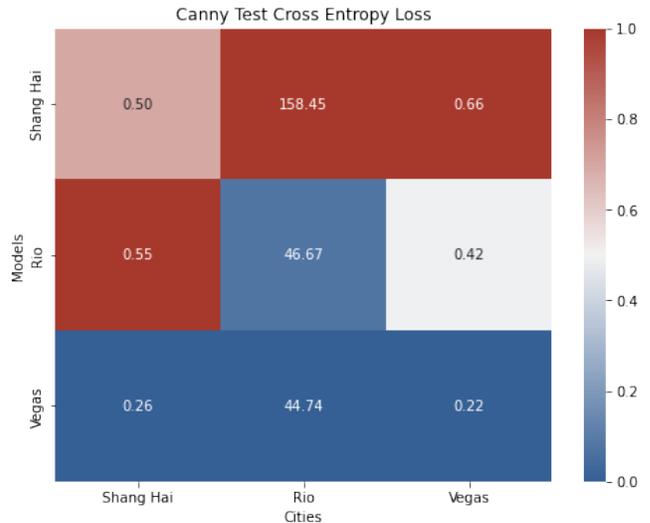

Figure 8: Categorical Inference Loss- lowest values on the diagonal indicate success

## 5. Distributional Transformation as a Defense Against Categorical Inference

### 5.1. Trojan Attack

We assess the attack's success by analyzing the Trojan accuracy and Benign accuracy under different poisoned data ratios and patch generation schemes. Recall our test data size of 1389. The baseline performance as measured on the test data is 0.93.

Tables 2 and 3 present the Trojan and Benign accuracy results, respectively, for various poisoned ratios (0.01, 0.1, 0.15, and 0.3). The poisoned ratio represents the proportion of training data that has been manipulated with adversarial patches. Our findings indicate that the Trojan attack is a very weak attack. Our best result for GAN based Trojans show it reducing the baseline performance by 11% with all other attacks being fairly ineffective. In the case of the PGD Trojan at 1% poisoned data the attack actually made the model perform *better*, indicating that at low levels of poisoning this attack may function as a regularization technique.

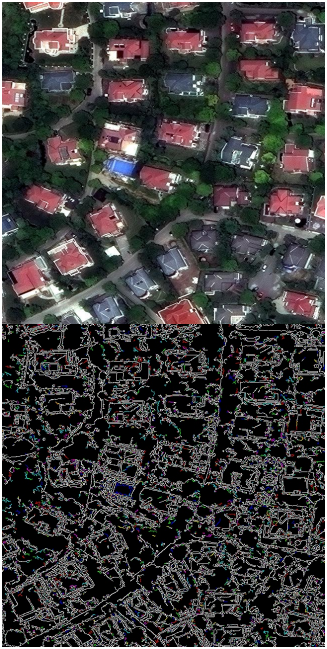

Figure 9: Canny Transformation of an image as a defense against Categorical inference from the Shanghai dataset. Note that we train on the Canny image but inference on the original image.

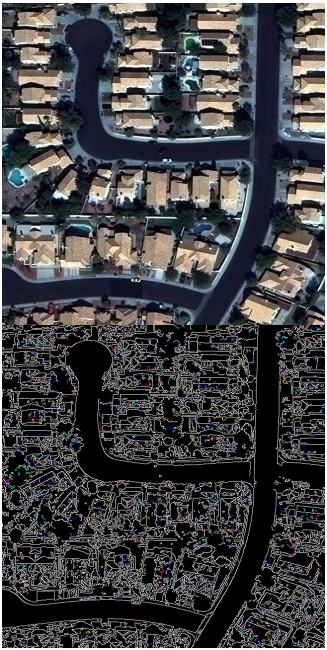

Figure 10: Canny Transformation of an image with buildings as a defense against Categorical inference from the Los Vegas dataset. Notice when compared to Shanghai the colors of roofs and the amount of green color in each image differ dramatically but are essentially uncorrelated to building presence. Canny removes this information.

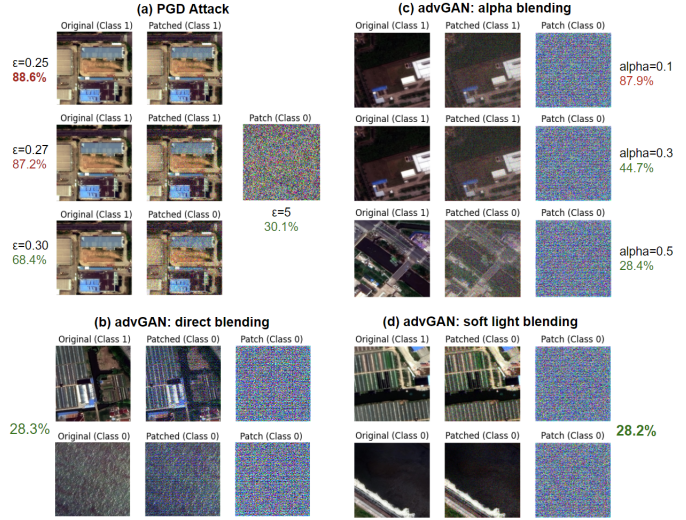

Figure 11: Patch Generation and Blending.

TABLE 2: Trojan Accuracy / Baseline Performance Accuracy (lower is better)

| Poisoned Ratio | PGD | GAN |
| --- | --- | --- |
| 1% | 101% | 89% |
| 10% | 99% | 99% |
| 15% | 97% | 99% |
| 30% | 97% | 96% |

TABLE 3: Benign Accuracy / Baseline Performance Accuracy (values close to 1 are best)

| Poisoned Ratio | PGD | GAN |
| --- | --- | --- |
| 1% | 98% | 98% |
| 10% | 98% | 98% |
| 15% | 98% | 97% |
| 30% | 95% | 97% |

## 6. Discussion and Conclusions

In this paper we have shown the relative ease of both evasion and categorical inference attacks and have shown that they apply in the overhead imagry domain. We have presented an in-depth exploration and analysis of adversarial patch blending techniques as well as their effects on segmentation and patch placement for adversarial attacks. We also conducted trojan attacks under various conditions; GAN-generated patches demonstrated moderate success rates with minimal poisoned data levels. These findings underscore the importance of carefully selecting the appropriate patch generation method and blending technique to maximize the effectiveness of adversarial attacks. The integration of building and water segmentation and patch placement into our pipeline allowed for more realistic patch placements, simulating physically placed patches. Our qualitative evaluations demonstrated the effectiveness of our segmentation models in accurately identifying building and water regions within images. This success highlights the potential of our pipeline to facilitate more realistic adversarial patch placements, contributing to an improved understanding of adversarial patch blending techniques and patch generation methods.

We have explored the efficacy of a trojan attack against a model trained on satellite images. In spite of extensive experimentation and exploration, our results indicate that the Trojan attack was not effective in our setting. We have identified three potential reasons for these negative results and discussed possible solutions.

***Hypothesis 1.*** *The trojan attack efficacy depends on the model being extremely accurate.*

Our experiments targeted a model that was merely good, but not close to perfect. The negative results add evidence that adversarial attacks against benchmark datasets may not directly transfer to real-world scenarios where models have lower accuracy. This observation aligns with prior work suggesting that the efficacy of poisoning attacks is dependent on model accuracy [33]. Future research could focus on refining the attack strategy in moderately powerful models or developing new attacks specifically designed for real-world scenarios.

***Hypothesis 2.*** *The trojan attack efficacy depends greatly on the patch being constant.*

In our experiments, we observed that even small variations in the trojan patch between training images could nullify the model's learned association. This behavior mirrors the exploitation of adversarial perturbations in other studies. Prior work has used patches with identical core structures, only translating or flipping them [34]. Future research could investigate the sensitivity of trojan attacks to patch variations and develop strategies to overcome this limitation.

***Hypothesis 3.*** *The trojan patch functioned as a confounder.*

In our study, the trojan patch was placed only in buildingless images, resulting in a high correlation between the patch and the buildingless feature data. This may have led the model to ignore the patch and focus on the building features instead. To address this issue, future work could consider alternative placement strategies for the trojan patch, ensuring it remains a strong signal for the model to learn without introducing confounding effects.

Despite the negative results observed in some of our experiments, our extensive exploration and thoughtful analysis provide a solid foundation for future research. Possible directions for future work include exploring the impact of different blending techniques on the robustness of adversarial attacks against various defense mechanisms, as well as investigating the applicability of our findings across diverse application domains. Additionally, researchers could focus on refining segmentation and patch placement methods to further enhance the realism and effectiveness of adversarial attacks.

# Appendix

All code can be found here in these two repositories: https://github.com/Lan131/adv_overhead/ and https://github.com/Lan131/adv_trojan

## F.1. Patch Generation

For evasion attacks we consider the following architectures:

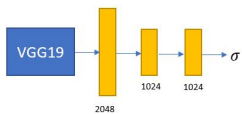

Figure 12: Rio Building Classifier architecture

Figure 13: Shanghai Building Classifier architecture

**F.1.1. PGD.** For adversarial patches, we varied the PGD hyperparmeter $\epsilon$ values (0.1, 0.2, 0.25, 0.272, and 0.3) to investigate the trade-off between the perturbation's magnitude and the model's prediction accuracy. Furthermore, we adjusted the step size $\alpha$ and the number of steps in our experiments to analyze their effect on the adversarial example (transparent) generation. The adversarial patches (opaque) are generated using the following parameters: $\epsilon = 5$ and $\alpha = 2/255$.

**F.1.2. advGAN.** For the advGAN patch generator, the input to the generator is a 100-dimensional noise vector, which is transformed into a 7x7x128 tensor by the fully connected layer. The transposed convolutional layers are responsible for upscaling the tensor to the desired image size (224x224) while progressively reducing the number of feature maps. The final output layer uses the tanh activation function to ensure the generated image pixel values are within the range of [-1, 1]. The discriminator is designed to take in both the real and generated images and output a probability indicating whether the input pair of images is real. The model has two convolutional layers, each followed by batch normalization and a leaky ReLU activation function. The output of the second convolutional layer is flattened and passed through a fully connected layer to produce a scalar value, which is then transformed into a probability using the sigmoid activation function [35]. The model architecture is summarized in Figures 14 and 15.

The GAN model was trained using the Adam optimizer with a learning rate of 0.0002 and beta values of 0.5 and 0.999 for both the generator and discriminator. The Binary Cross-Entropy (BCE) loss function was used as the objective function for both components. The training process involved alternating between updating the discriminator and the generator. For each mini-batch, the discriminator was trained by comparing its predictions for both real and generated images against the true labels (1 for real images and 0 for fake images). The generator was then updated to minimize the BCE loss between the discriminator's predictions for the generated images and the real labels (1), effectively making the generated images more difficult to discern from real ones. The training process was carried out for five epochs, and the generator and discriminator losses, as well as the real and fake accuracy, were recorded for each epoch [35].

```
----------------------------------------------------------------
        Layer (type)               Output Shape         Param #
================================================================
            Linear-1                  [-1, 6272]         633,472
   ConvTranspose2d-2          [-1, 64, 14, 14]          131,136
       BatchNorm2d-3          [-1, 64, 14, 14]              128
   ConvTranspose2d-4          [-1, 32, 28, 28]           32,800
       BatchNorm2d-5          [-1, 32, 28, 28]               64
   ConvTranspose2d-6          [-1, 16, 56, 56]            8,208
       BatchNorm2d-7          [-1, 16, 56, 56]               32
   ConvTranspose2d-8          [-1, 8, 112, 112]           2,056
       BatchNorm2d-9          [-1, 8, 112, 112]              16
  ConvTranspose2d-10         [-1, 3, 224, 224]              387
================================================================
Total params: 808,299
Trainable params: 808,299
Non-trainable params: 0
```

Figure 14: Generator Model.

```
----------------------------------------------------------------
        Layer (type)               Output Shape         Param #
================================================================
            Conv2d-1         [-1, 64, 112, 112]           6,208
       BatchNorm2d-2         [-1, 64, 112, 112]             128
            Conv2d-3         [-1, 128, 56, 56]          131,200
       BatchNorm2d-4         [-1, 128, 56, 56]              256
            Linear-5                   [-1, 1]          401,409
     Discriminator-6                   [-1, 1]                0
================================================================
Total params: 539,201
Trainable params: 539,201
Non-trainable params: 0
```

Figure 15: Discriminator Model.

To generate patches, the generator is fed with random noise, which it then transforms into a patch-like structure using its learned parameters. The adversarial patches are created by the generator to fool the target model. The generated patches are combined with the original data, and the discriminator is trained to classify whether the combination of original and adversarial data is genuine or manipulated. The generator's objective is to create patches that can deceive the discriminator, while the discriminator's goal is to accurately classify the combined data. Through this adversarial process, the generator becomes increasingly adept at creating patches that can successfully fool the target model. The attack's performance was evaluated by comparing the target model's accuracy on original images and adversarial images generated by different blending techniques.

## F.2. Blending Details

**F.2.1. Direct Blending.** In the direct blending method, adversarial patches are directly overlaid onto the original images without considering the pixel values of the original image. This approach creates a simple combination of both images, where the adversarial patch completely replaces the original image's pixel values in the region it covers. Although this method is straightforward, it may not always produce visually consistent or smooth results, as the transition between the original image and the adversarial patch can be abrupt and noticeable. However, direct blending is computationally efficient and can be easily implemented, making it a suitable option for preliminary evaluations of adversarial attacks.

**F.2.2. Alpha Blending.** Alpha blending is an advanced technique used to merge an original image with its adversarial patch by assigning an alpha value (or weight) to each image's pixel values, creating an effortless transition. The alpha parameter ranges from 0-1 and defines the level of blending: setting it at zero will leave the original image undisturbed, while 1 will completely replace it with its adversarial patch. Adjusting this parameter enables users to optimize attack effectiveness vs patch visibility trade-offs; hence making Alpha blending an attractive choice used across computer vision applications.

**F.2.3. Soft Light Blending.** Soft light blending is an advanced blending technique that emulates the effect of shining diffuse light onto an original image by using its adversarial patch as a light source. This method results in a more visually coherent integration of the adversarial patch and the original image, allowing the patch to appear as a natural part of the scene. Soft light blending combines the original image and the adversarial patch by applying a mathematical function that models the behavior of light, providing a more realistic appearance. Although soft light blending may be more computationally expensive than direct blending or alpha blending, its ability to produce visually consistent results makes it an attractive option for applications where the adversarial patch needs to appear inconspicuous or seamlessly integrated into the original image [36].

### F.3. CNN Models to Attack

In this study, we utilized the ResNet 152 model with batch normalization as our target model for the adversarial attacks. VGG-19 is a well-established deep convolutional neural network architecture that has demonstrated excellent performance in image classification tasks. It consists of 19 layers with learnable weights, including 16 convolutional layers and 3 fully connected layers. The model has been pretrained on the ImageNet dataset, which comprises over 1.2 million images spanning 1000 classes [37].

To adapt the VGG19 model for our specific binary classification task, we replaced the original classifier with a custom classifier. This new classifier consists of a series of linear layers, Leaky ReLU activation functions, and dropout layers, ultimately producing two output classes. The final output layer employs a LogSoftmax function, which normalizes the output probabilities and allows for easier interpretation of the results. To prevent the model's pretrained features from being altered during the training process, we froze all the parameters in the VGG19 model by setting the requires_grad attribute to False. This ensures that only the newly added custom classifier's weights are updated during training, leveraging the pretrained features for our specific task without modifying them. The training process involved using a loss function, an optimizer, and a specified number of epochs. In this case, we trained the model for 10 epochs. By monitoring the model's performance on the validation set, we were able to track the learning progress and avoid overfitting. Using the fine-tuned VGG19 model as our target, we proceeded with the adversarial attacks, investigating the vulnerability of this architecture to adversarial examples and assessing the effectiveness of different attack parameters.

### F.4. Patch Blending Ablations

Our preliminary findings, based on training with the simple MNIST dataset, reveal that GAN-generated patched images can deceive a CNN model, as demonstrated by the accuracy on original images (0.9911) compared to the accuracy on adversarial images (0.3205). However, as seen in Figure 16, the patches do not seamlessly blend into the images, making them easily discernible. To address this limitation learned from the baseline Mnist dataset, our next step focused on investigating novel patch blending techniques that produce more natural-looking images.

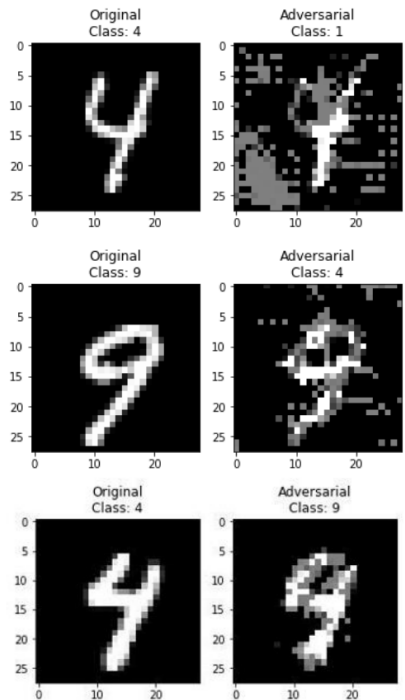

Figure 16: GAN Patch Mnist Visualization.

We conducted three experiments to assess the effectiveness and visual consistency of direct blending, alpha blending, and soft light blending techniques in generating adversarial patches. In each experiment, we employed the same adversarial patch generation method (direct blending for PGD patches due to its inherent blending nature and all three techniques for advGAN) and used the pre-trained VGG deep learning model for image classification. We evaluated the classification accuracy on both original and patched images and examined the qualitative appearance of the patched images. Furthermore, we evaluated the performance of three blending techniques under differing conditions to gain more insight into their strengths and weaknesses. Figure 11 presents our experiments' results. The VGG-19 model's classification test accuracy was approximately 90%. For the attacks, the green-colored accuracies represent successful attacks, while red indicates unsuccessful attacks. Columns one, two, and three in Figure 11-(a) to -(d) display the original, patched images, and patches, respectively. Class labels are provided at the top of the images. Figure 11-(a) presents the baseline PGD attack accuracies for different epsilon values, with accuracy directly proportional to visual detectability, as well as the patch generated using PGD.

In the first experiment (Figure 11-(a) and -(c) inset), we applied direct blending by overlaying the adversarial patches onto the original images. The inset shows the patch used for direct blending, with the overlaid results omitted since they closely resemble the middle column in (c). The quantitative results revealed a significant drop in classification accuracy for patched images compared to the original images, indicating the adversarial attack's effectiveness when using direct blending. However, the qualitative analysis showed that adversarial patches were highly visible and created abrupt transitions, making the attack more noticeable and less visually consistent. These abrupt transitions may be problematic for adversarial attacks requiring a subtle or inconspicuous appearance. The second experiment employed alpha blending with varying degrees of blending strength (alpha = 0.1, 0.3, and 0.5). Our quantitative analysis demonstrated that as the alpha value increased, classification accuracy for the patched images decreased, indicating the adversarial attack's increased effectiveness. This result suggests that alpha blending can strike a balance between attack effectiveness and visual appearance by adjusting the alpha value. However, higher alpha values also led to more visible patches, revealing a trade-off between the attack's effectiveness and the patch's visibility. Qualitatively, alpha blending produced smoother transitions between the original images and the adversarial patches, resulting in a more visually appealing combination. This improvement in visual appearance could make the attack more challenging to detect by human observers. In the third experiment, we implemented soft light blending to combine the original images and the adversarial patches. Quantitative results showed that the soft light blending approach was effective in decreasing the classification accuracy of the patched images, comparable to the results obtained with alpha blending. These results indicate that soft light blending can provide a high level of adversarial attack effectiveness. However, the qualitative analysis revealed that the soft light blending technique produced the most visually consistent and natural-looking patched images.

The adversarial patches appeared seamlessly integrated into the original images, making the attack less noticeable and more challenging to detect. This seamless integration could be particularly advantageous for adversarial attacks that need to evade both human and machine detection.

Our experiments demonstrated that while all three blending techniques were effective in reducing classification accuracy, soft light blending provided the most visually consistent and natural appearance. This finding suggests that soft light blending is the preferred technique when implementing adversarial attacks requiring seamless integration of the adversarial patch into the original image. Future research could explore the impact of different blending techniques on the robustness of adversarial attacks against various defense mechanisms and in diverse application domains [38].

The models were trained on two datasets, the Shanghai dataset and Mnist dataset. The Shanghai dataset contained 3,650 training images and 927 test images; Mnist contained 60,000 images with an 8:2 train-to-test ratio. The models trained on the Shanghai dataset showed an accuracy of 91% on the test set (99% for Mnist), including initial models used as attack vectors for PGD attacks as well as models trained with poisoned data. From a representative example shown in Figure 2, we can see after four epochs, the training and validation losses converged to around 0.25 and 0.20, respectively, signifying that our models could generalize well without overfitting. Accuracy on both datasets (Shanghai and Mnist) provides further evidence that our models are adept at handling real-world data, and serves as the basis for future experiments involving adversarial attacks. Such attacks will assess various blending techniques, adversarial patch generation methods, and patch placement strategies within the adversarial machine learning context.

To evaluate the effectiveness of the three blending techniques (direct blending, alpha blending, and soft light blending), we conducted a series of experiments using the pre-trained VGG deep learning model for image classification. In each experiment, we measured the classification accuracy for both benign (non-trojan) and trojan (patched) images. The benign accuracy refers to the model's ability to correctly classify unaltered images, while trojan accuracy represents the model's ability to misclassify images with adversarial patches.

### F.5. Segmentation and Patch Placement Results

The effectiveness of our segmentation and patch placement pipeline was assessed through qualitative evaluations, as ground truth segmentation masks were not available for our dataset. The results demonstrated high accuracy for both building and water segmentation.

**F.5.1. Building Segmentation Results.** Qualitatively, our building segmentation model performed well on a wide range of images, accurately detecting polygons in most cases. Although a quantitative evaluation could not be conducted due to the lack of ground truth segmentation masks, the visual assessment suggested that the model was highly effective. Figure 17 presents several examples of successful building segmentation.

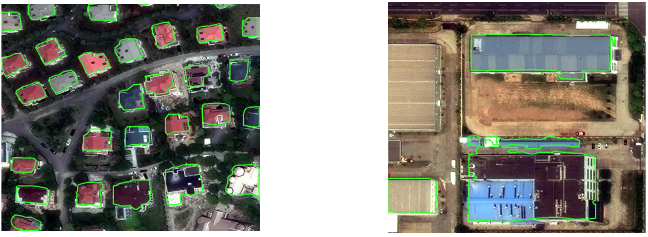

Figure 17: Building Segmentation Examples

**F.5.2. Water Segmentation.** The water segmentation results are illustrated in Figure 18.

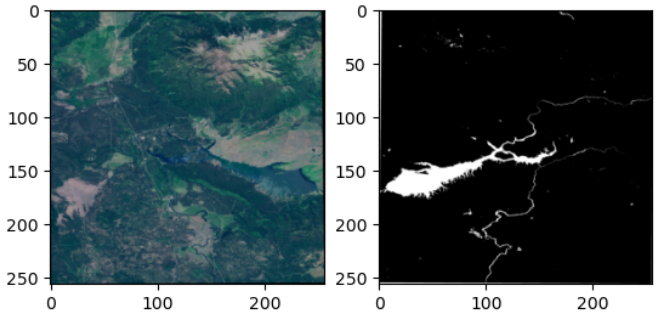

Figure 18: Water Segmentation

We trained on [39] setting the ratio threshold for determining whether an image contains water at 0.5. To evaluate the accuracy of the segmentation method, we randomly selected 50 ocean images and found that 92% of them were correctly classified. This high classification rate indicates the effectiveness of our water segmentation approach.

# References

TABLE 4: Model Summary

| Layer (type) | Output Shape | Param # |
|---|---|---|
| Conv2d-1 | [-1, 64, 128, 128] | 1792 |
| BatchNorm2d-2 | [-1, 64, 128, 128] | 128 |
| ReLU-3 | [-1, 64, 128, 128] | 0 |
| Conv2d-4 | [-1, 128, 64, 64] | 73856 |
| BatchNorm2d-5 | [-1, 128, 64, 64] | 256 |
| ReLU-6 | [-1, 128, 64, 64] | 0 |
| Conv2d-7 | [-1, 256, 32, 32] | 295168 |
| BatchNorm2d-8 | [-1, 256, 32, 32] | 512 |
| ReLU-9 | [-1, 256, 32, 32] | 0 |
| Conv2d-10 | [-1, 512, 16, 16] | 1180160 |
| BatchNorm2d-11 | [-1, 512, 16, 16] | 1024 |
| ReLU-12 | [-1, 512, 16, 16] | 0 |
| Conv2d-13 | [-1, 1024, 8, 8] | 4719616 |
| BatchNorm2d-14 | [-1, 1024, 8, 8] | 2048 |
| ReLU-15 | [-1, 1024, 8, 8] | 0 |
| ConvTranspose2d-16 | [-1, 512, 16, 16] | 2097664 |
| BatchNorm2d-17 | [-1, 512, 16, 16] | 1024 |
| ReLU-18 | [-1, 512, 16, 16] | 0 |
| ConvTranspose2d-19 | [-1, 256, 32, 32] | 1179904 |
| BatchNorm2d-20 | [-1, 256, 32, 32] | 512 |
| ReLU-21 | [-1, 256, 32, 32] | 0 |
| ConvTranspose2d-22 | [-1, 128, 64, 64] | 295040 |
| BatchNorm2d-23 | [-1, 128, 64, 64] | 256 |
| ReLU-24 | [-1, 128, 64, 64] | 0 |
| ConvTranspose2d-25 | [-1, 64, 128, 128] | 73792 |
| BatchNorm2d-26 | [-1, 64, 128, 128] | 128 |
| ReLU-27 | [-1, 64, 128, 128] | 0 |
| Conv2d-28 | [-1, 1, 128, 128] | 65 |
| Total params: | 8,576,833 | |
| Trainable params: | 8,576,833 | |
| Non-trainable params: | 0 | |